\newcommand{\includesvg}[1]{
}

\documentclass[10pt,twocolumn]{article}

\pdfoutput=1

\usepackage{cvpr}
\usepackage{times}
\usepackage{graphicx}
\usepackage{amsmath}
\usepackage{amssymb}
\usepackage[farskip=5pt]{subfig}
\usepackage{afterpage}
\usepackage{cite}

\usepackage[colorlinks,bookmarks=false]{hyperref}

\cvprfinalcopy 

\def\cvprPaperID{119} 

\ifcvprfinal\fi

\begin{document}
\title{Difference of Normals as a Multi-Scale Operator in Unorganized Point Clouds}
\author{Yani Ioannou\textsuperscript{1}\\
{\tt\small yani@cs.utoronto.ca}
\and
Babak Taati\textsuperscript{1}\\
{\tt\small taati@cs.utoronto.ca}
\and
Robin Harrap\textsuperscript{2}\\
{\tt\small harrap@geol.queensu.ca}
\and
Michael Greenspan\textsuperscript{3}\\
{\tt\small michael.greenspan@queensu.ca}
\and
{\textsuperscript{1} University of Toronto\,/\,Toronto Rehabilitation Institute, Toronto, Ontario, Canada }\\
{\textsuperscript{2} Queen's GIS Laboratory, Queen's University, Kingston, Ontario, Canada}\\
{\textsuperscript{3} Dept. of Electrical \& Computer Engineering, Queen's University, Kingston, Ontario, Canada}
\thanks{\tiny \copyright 2012 IEEE. Personal use of this material is permitted. Permission from IEEE must be obtained for all other uses, in any current or future media, including reprinting/republishing this material for advertising or promotional purposes, creating new collective works, for resale or redistribution to servers or lists, or reuse of any copyrighted component of this work in other works.}
}

\maketitle
\thispagestyle{empty} 
\begin{abstract}
A novel multi-scale operator for unorganized 3D point clouds is introduced. The Difference of Normals (DoN) provides a computationally efficient, multi-scale approach to processing large unorganized 3D point clouds. The application of DoN in the multi-scale filtering of two different real-world outdoor urban LIDAR scene datasets is quantitatively and qualitatively demonstrated. In both datasets the DoN operator is shown to segment large 3D point clouds into scale-salient clusters, such as cars, people, and lamp posts towards applications in semi-automatic annotation, and as a pre-processing step in automatic object recognition. The application of the operator to segmentation is evaluated on a large public dataset of outdoor LIDAR scenes with ground truth annotations.
\end{abstract}


\section{Introduction}

\subsection{Motivation}
The increasing prevalence of 3D scanners has resulted in a dramatic explosion in the availability of 3D data, especially raw sensor data often represented in the most basic 3D point cloud format. Such sensors include LIDAR scanners for modelling large outdoor scenes and GIS applications, as well as commercially available and inexpensive solutions for indoor scanning and modelling. Consequently the processing of point clouds of millions, or even hundreds of millions of points has become commonplace. Furthermore, new applications of range sensors that require the processing of large point clouds in real-time, such as self-driving cars~\cite{KITTI}, have arose.

For such datasets and their applications to be useful, or even feasible, there is a demand for salient point selection algorithms based solely on an unorganized point cloud - as opposed to connected-graph and mesh-based algorithms which are typically more computationally and memory intensive. This has motivated the move towards using simple point cloud processing algorithms to filter a point cloud for salient points before applying complex algorithms, akin to the common usage of image processing filters in 2D computer vision algorithms.

One such image processing filter is the Difference of Gaussians (DoG). The DoG is an approximation to the Laplacian of the Gaussian (LoG) operator, and is widely used in applications such as image enhancement, blob detection, edge detection, finding points of salience, pre-segmenting images~\cite{maar_hildreth}, and perhaps most notably in the form of a DoG pyramid for obtaining scale invariance in 2D object recognition~\cite{SIFT}. Although the DoG operator easily generalizes to so-called 2.5D data (i.e.\ depth images) and volumetric images (e.g.\ in medical imaging)~\cite{volumetricscale}, extending it to unorganized data (i.e.\ point clouds), particularly in a computationally efficient manner, is less straight forward.

\subsection{Contributions}
\label{sec:contributions}
In this paper, a multi-scale operator of similar function to the DoG is introduced for unorganized 3D point clouds, namely the Difference of Normals (DoN). Despite the simplicity and efficiency of the operator, DoN is shown to be surprisingly powerful in assigning point saliency according to scale. While DoN is motivated in this paper as a multi-scale saliency feature used in a segmentation and/or object recognition pipeline, it also has applications to oriented 3D edge detection and planar region segmentation. An open source implementation of the DoN operator is made available in the Point Cloud Library (PCL)~\cite{PCL} \footnote{Available as a feature in the PCL trunk: \href{http://www.pointclouds.org}{http://www.pointclouds.org}.}. 

\subsection{Organization}
\label{sec:organization}
The rest of this paper is organized as follows:
In \S\ref{sec:prevwork} previous work on multi-scale operators and segmentation in unorganized point clouds is summarized;
\S\ref{sec:donoperator} introduces the Difference of Normals (DoN) operator;
\S\ref{sec:results} introduces a method for parameter selection, and shows the application of the DoN operator to the segmentation of real urban LIDAR scenes. A quantitative analysis of DoN segmentations from a publicly available dataset with ground truth annotations is evaluated;
\S\ref{sec:conclusion} summarizes the results and explores the potential for future work.

\section{Previous Work}
\label{sec:prevwork}
\subsection{Scale and Unorganized Point Clouds}
In 2D images, the concept of scale space is often described with a family of gradually smoothed images created through the convolution of a Gaussian kernel, such a Gaussian scale-space has a wide range of applications in image processing and 2D computer vision, such as in edge sharpening and interest point selection~\cite{SIFT}. 

Extending the concept of scale-space to unorganized 3D point clouds is a non-trivial task due to the lack of a regular lattice from which points were sampled. One solution is to convert the data into an organized format, such as a dense voxel map, in which the generalization from 2D image processing is straightforward. This however, is an unrealistic task for large point clouds, such as outdoor scenes with millions of points, since the required number of voxels will be vast. Octree representation is a possible solution in reducing the memory requirements, but it comes at a computational cost. 

Unnikrishnan et al., arguing the need for a multi-scale unorganized point cloud operator, introduced such an operator derived from Laplace-Beltrami operator - a generalization of the Laplacian to Riemannian manifolds~\cite{Unnikrishnan_2008}.  Their proposal of a multi-scale unorganized point cloud operator was derived in a scale-space theoretic method, and being based on a Gaussian convolution kernel satisfies the scale space axioms. In application, however, the operator is relatively computationally and memory intensive compared with our proposed method, as it requires the computation of a geodesic distance graph for the point cloud and the convolution kernel is computationally expensive to compute. Furthermore it is unclear how the operator could be used to detect oriented features such as corners or edges.

\subsection{Normal Support Radius in 3D Point Features}
Many proposed features for unorganized point clouds have directly, or indirectly used the relation of support region size in surface normal estimation as a method of scale or saliency detection on the implicit surfaces of a 3D point clouds. 

Rusu et al.\ proposed Persistent Point Feature Histograms for unorganized point clouds. These features were calculated in part using a series of normals estimated with a series of increasing radii between a fixed minimum and maximum radius~\cite{Rusu_Marton_Blodow_Beetz_2008}. Novatnack and Nishino used surface normals to create scale-dependent geometric features on triangular meshes. The mesh normals were interpolated over a 2D parametrization of the mesh over each vertex, creating a normal map that was used in edge detection. They argued that while a normal map does not satisfy the axioms of a scale space operator, it is a natural choice as normals are less effected by noise as compared to higher order derivative quantities such as curvature.

\subsection{Point Cloud Segmentation}
Various algorithms have been proposed in the area of point cloud segmentation. However, most algorithms for unorganized point clouds have require meshing or connectivity~\cite{Huang, Golovinskiy_Funkhouser_2009}. Those that do not often require estimating the normal map as an integral step.

Liu et al.\ introduced Cell Mean Shift (CMS)~\cite{cms}, which maps the normal map of a point cloud to a Gaussian sphere, producing a \emph{Gaussian image}. This spherical image can then be clustered to identify shapes. Woo et al.~\cite{woo} propose an octree-based method for handling large  point clouds, using edges to segment structures within. 

\section{Difference of Normals Operator}
\label{sec:donoperator}
\subsection{Theory}
\label{sec:theory}
The concept of scale-space has a well established theoretical background for continuous and discrete signals, notably in linking the relationship between scale-space and the linear diffusion equation~\cite{koenderink} and in establishing the scale space axioms~\cite{koenderink,witkin}. A set of axioms~\cite{koenderink,scalespacebook,witkin}, the complete review of which is beyond the scope of this paper, are described to capture the properties of a desired and useful scale-space representation. Notably, it has been proven that the Gaussian kernel is the only convolutional filter that satisfies the complete set of scale-space axioms~\cite{koenderink}. 

Although the Gaussian kernel is unique in satisfying the complete set of scale-space axioms, for many applications an operator that satisfies the full set of scale-space axioms is not required. Such approaches are more generally referred to as \emph{multi-scale}. Multi-scale operators may be simpler, more computationally efficient and have desirable properties such as orientability.

This paper proposes to define a multi-scale operator for unorganized point clouds directly using the estimated surface normal map of an unorganized point cloud. The primary motivation behind this, is the observation that surface normals estimated at any given radius reflect the underlying geometry of the surface at the scale of the support radius. Although there are many different methods of estimating the surface normals (see \S\ref{sec:normalestimation}), normals are always estimated with a support radius (or via a fixed number of neighbours). This support radius determines the scale in the surface structure which the normal represents. 

\begin{figure}[htb]
	\centering
	\includegraphics[width=.95\linewidth]{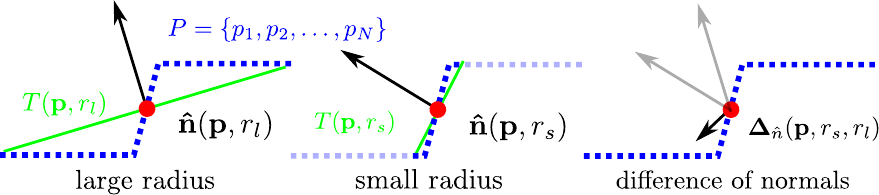}
	\caption{The normal support radius' relation to scale.}
	\label{fig:normalsupporteffect}
\end{figure}

Fig.~\ref{fig:normalsupporteffect} illustrates this effect in 1D. Normals, $\mathbf{\hat{n}}$, and tangents, $T$, estimated with a small support radius $r_s$ are affected by small-scale surface structure (and similarly by noise). On the other hand, normals and tangent planes estimated with a large support radius $r_l$ are less affected by small-scale structure, and represent the geometry of larger scale surface structures. 

	The intuition behind the approach being proposed here is that if the direction of the two surface normals is nearly identical, then the structure of the surface does not change significantly from the first radius to the second. By contrast, if the structure of the larger neighbourhood around a center point  is significantly different from that of the smaller neighborhood, then the direction of the two estimated normals are likely to vary by a larger margin. In that case, a value between the two radii is often a representative  of the scale around near the center point. 
 
	Suppose a multi-scale operator for a point cloud is simply defined as:
	\begin{align}
		\mathbf{L}(\mathbf{p}, r) &= \mathbf{\hat{n}}(\mathbf{p}, r),\label{normalscaleoperatoreqn}
	\end{align}
	with scale parameter $r$, effected by the normal map of a point cloud $P$ estimated with support radius $r$. Notice the response of our operator is a vector, and is thus orientable, however the operator's $l_2$ norm provides a more conventional scale quantity. 

	Just as described by the most basic and intuitive scale space axiom, the effect of the normals on the implicit surface sampled by a point cloud is to suppress most of the structures in the surface with a characteristic dimension of less than $r$. Furthermore, with increasing values of the scale parameter $r$, fine scale surface structure is increasingly suppressed. Despite this, Eqn.~\ref{normalscaleoperatoreqn} does not satisfy all scale space axioms originally outlined by Witkin et al.~\cite{witkin} and more recently enumerated by Lindeburg et al.~\cite{scalespacebook}. Notably the causality requirement introduced by Koenderink et al.~\cite{koenderink}.

\subsection{Method}
\label{sec:method}
	When applying the multi-scale operator defined in Eqn.~\ref{normalscaleoperatoreqn}, we compare the responses at each point $\mathbf{p}$ over several radii $r_1 < r_2 < \ldots < r_n$. In the most basic case we can compare the response of the operator across two different radii $r_1<r_2$. Formally, the Difference of Normals (DoN) operator $\mathbf{\Delta}_\mathbf{\hat{n}}$  for any point $\mathbf{p}$ in a point cloud $P$, is defined as:
	\begin{align}
		\mathbf{\Delta}_\mathbf{\hat{n}}(\mathbf{p}, r_1, r_2) =& \frac{\mathbf{\hat{n}}(\mathbf{p}, r_1) - \mathbf{\hat{n}}(\mathbf{p}, r_2)}{2},
	\end{align}
\noindent where  $r_1, r_2 \in \mathbb{R}$, $r_1<r_2$, and $\mathbf{\hat{n}}(\mathbf{p}, r)$ is the surface normal estimate at point $\mathbf{p}$, given the support radius $r$.

	For a given $r_1$ and $r_2$, the result of applying the $\mathbf{\Delta}_\mathbf{\hat{n}}$ operator to all the points in a point cloud is a vector map where a DoN vector is assigned to each point.  Since each DoN is the normalized sum of two unit normal vectors, the magnitude of the $\mathbf{\Delta}_\mathbf{\hat{n}}$ vectors are always within $\left[ 0, 1 \right]$. 

The DoN vectors may be thresholded based on their magnitude, i.e.\ $\| \mathbf{\Delta}_\mathbf{\hat{n}}(\mathbf{p}) \|$, or based on their component values, i.e.\ $\Delta_\mathbf{\hat{n}}^x(\mathbf{p})$, $\Delta_\mathbf{\hat{n}}^y(\mathbf{p})$, or $\Delta_\mathbf{\hat{n}}^z(\mathbf{p})$ for orientable surfaces and edges. 

	Calculating the two normal maps estimated with support radii $r_1, r_2$ for a scene is a process which is highly parallelizable and thus greatly benefits from GPU optimization. Consequently, DoN computation, even for very large scale point clouds, may be performed very efficiently (see \S\ref{subsec:computation}).

\subsection{Approximating Normals in Range Data}
\label{sec:normalestimation}
\subsubsection{Normals Estimation}
	There are many methods for estimating normals (or equivalently tangent planes) in point clouds~\cite{hoppenormals,huinormals,alexanormals}. However, only those using a fixed support radius, rather than a fixed number of neighbors, are suitable for unorganized data, especially when the point cloud density is highly variable. 

	Applying a method based on a fixed number of neighbors to a point cloud with a high variability in sampling density, e.g.\ urban LIDAR data, results in each normal being computed using what may be a very different support radii, and thus the estimated normals at each point will represent the surface at very different scales. Such normals would be unsuitable for DoN calculations.

	In our experiments, the normals were estimated by finding the tangent plane using the principal components of a local neighborhood of fixed support radius around each point. This neighborhood may contain any number of points $N\ge 3$. The result is that all the normals in the scene are calculated at the same scale. However, due to the highly variable sampling density/resolution of some range data, the accuracy of the normal estimate, may vary considerably across a scene with $N$. 

	It is important to note that PCA is not robust to outliers, and for some applications more robust methods of normal estimation may be more suitable~\cite{robustnormals}, however in our experiments we found PCA estimated normals to be sufficient even in the presence of highly unorganized data.

\subsubsection{Resolving Normal Ambiguity}
	Surface normals estimated on point clouds exhibit a sign ambiguity in their direction. This is because any tangent plane to a point has two normals in opposite directions, either of which is mathematically valid. In many applications, this normal ambiguity is typically resolved with the sensor context, since the correct normal is always the one pointing in the hemisphere towards the range sensor~\cite{taati}. 
	For the particular application of DoN operations, the particular choice of resolving the normal sign ambiguity has no consequence so long as the normals for the two support radii are disambiguated in the same manner. Thus the disambiguation of the normals can simply be achieved by negating one of the normals if $\mathbf{\hat{n}}(\mathbf{p}, r_1) \cdot \mathbf{\hat{n}}(\mathbf{p}, r_2) > \frac{\pi}{2} $, i.e.\ if the angle between the two normals is greater than $90^\circ$. This is under the assumption that the true surface normals must be within an angle of $\frac{\pi}{2}$ of each other, which is a realistic assumption given the limitation of scanning in the presence of self-occlusion. 
\section{Experimental Results}
\label{sec:results}

\subsection{Parameter Selection}
\label{sec:parameterselection}
Selecting the parameters $r_1$ and $r_2$ for DoN is important since, while a wide range of parameters may elicit large responses from the surface of interest, naive parameters choices may also have large responses in other classes of surfaces. We propose a simple parameter selection algorithm where the objective is to choose parameters maximizing the DoN magnitude for the set of points within the objective class, while minimizing the DoN magnitude for other known classes of surfaces/objects in the scene.

In practice, given a set of ground truth point clouds for our objective object (e.g.\ cars) and a set of ground truth point clouds for the objects in close vicinity of the objective object (e.g. road, people), we compare the aggregate response statistics (i.e.\ median, mean, variance) for all points in each class across a selection of DoN parameters. 

\begin{figure}[htbp]
	\centering
        \subfloat[Car aggregate statistics]{
        	\includegraphics[width=0.95\linewidth,keepaspectratio]{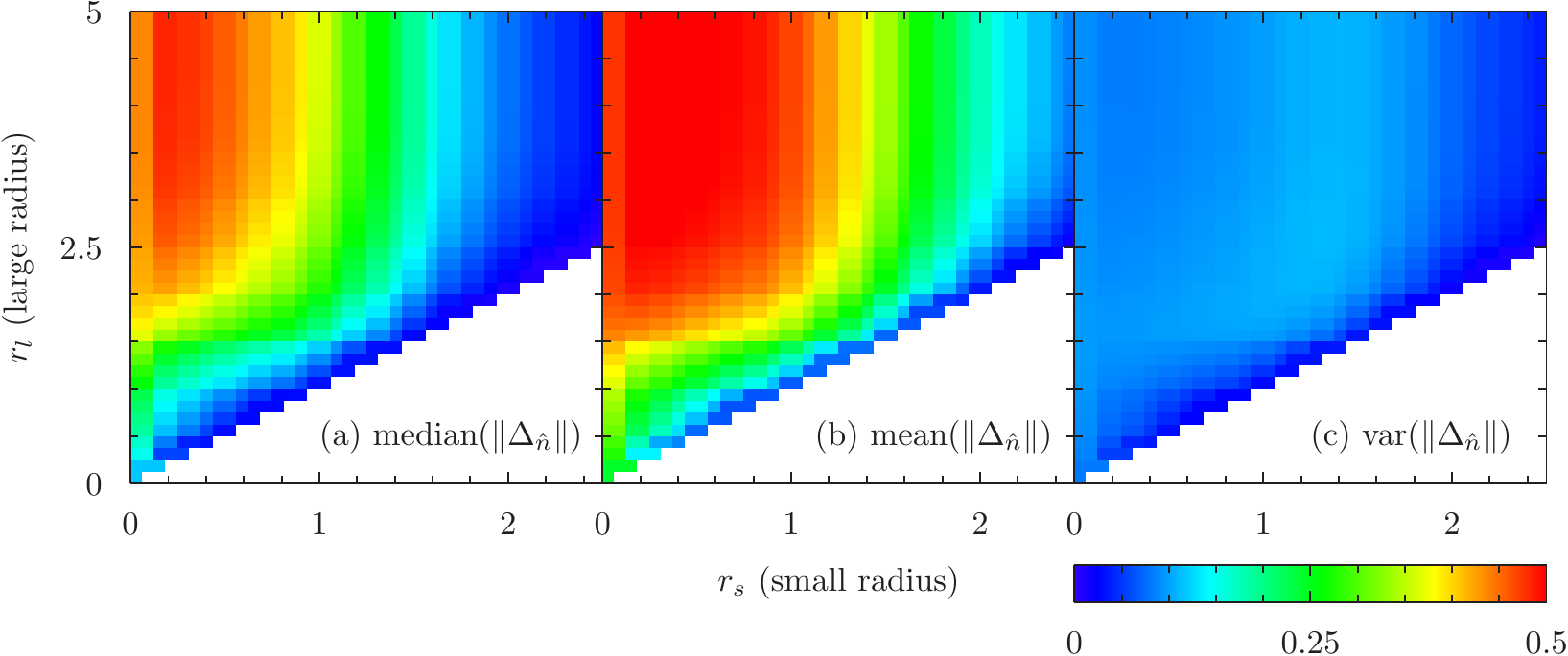}
                \label{fig:car_parammap}
        }\\
        \subfloat[Pedestrian aggregate statistics]{
        	\includegraphics[width=0.95\linewidth,keepaspectratio]{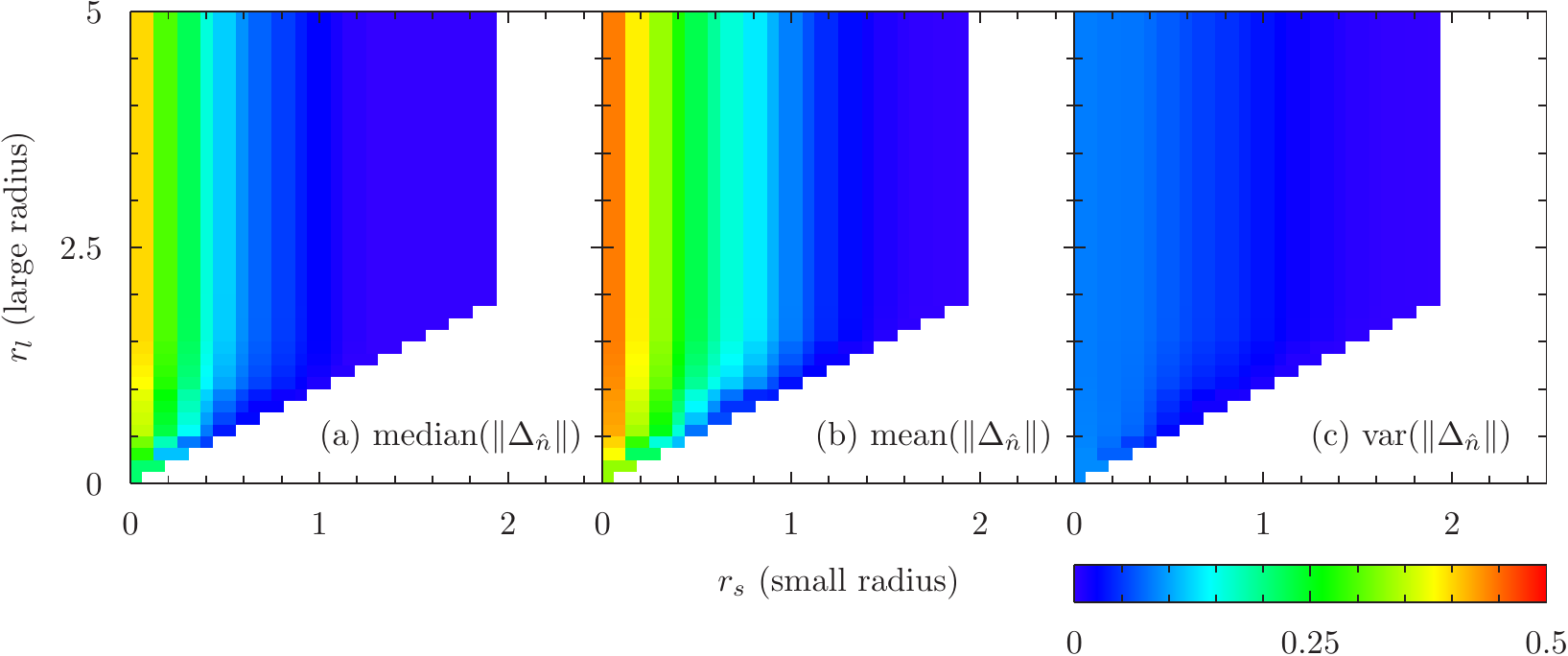}
                \label{fig:pedestrian_parammap}
        }
	\caption[]{Aggregate per-object class statistics used in parameter selection.}
        \label{fig:paramcolourmaps}
\end{figure}

Fig.~\ref{fig:paramcolourmaps} shows the mean, median and variance responses for a set of object classes from a single data sequence in the KITTI dataset~\cite{KITTI} over a range of parameters $r_s, r_l$. Using this, for example, we empirically set the parameters $r_s=0.1, r_l=0.4$ for pedestrians and $r_s=0.4, r_l=2.0$ for cars in order to maximize the intra-class response distance in a scene containing both objects in close proximity.

\subsection{TITAN Urban Mobile LIDAR Data}
\label{sec:titanresults}
	In GIS applications, there is a focus on the recognition of street furniture (a GIS term describing lamp posts, fire hydrants, curbs, etc.) and the extraction of large-scale infrastructure (e.g.\ buildings, roads). The DoN operator is an ideal tool for addressing such problems by isolating objects in the scene based on their scale. 
		The following results demonstrate various applications of DoN to real-world data, towards automatic segmentation and as a pre-processing step in object recognition or annotation in urban LIDAR data.
		
		The points clouds used in the experiments reported here are from a TITAN system~\cite{titan}.
		The raw data is collected via a LIDAR scanner mounted on a moving vehicle and scanning the urban scenery as the vehicle traverses the street. The individual scans are then registered together to form complete 3D point clouds of large urban areas.  
			As the registration of many low ($\sim0.1$~cm) resolution scans from a mobile platform, the final point clouds have a highly variable sampling density governed mostly by the speed of the vehicle, the changing obstructions in the scene, and the registration error. Despite the large amount of data in a typical TITAN scene, individual objects are often composed of small numbers of points, e.g.\ $\sim100$ points in the case of a person. All these factors make processing the TITAN point clouds a particularly challenging domain.

	For illustration purposes, the results in this section are demonstrated using small ($25$ $\textrm{m}^3$) sections of a real-world urban LIDAR data in the city of Kingston, ON, Canada, collected by the TITAN mobile terrestrial scanner. Similar results were also observed on much larger datasets of hundreds of millions of points. 

	\subsubsection{DoN Features}
	\label{subsec:donoutput}
		\begin{figure}[htbp]
		\centering
		\subfloat[Point cloud (478,377 points).]{
			\includegraphics[width=.45\linewidth,keepaspectratio]{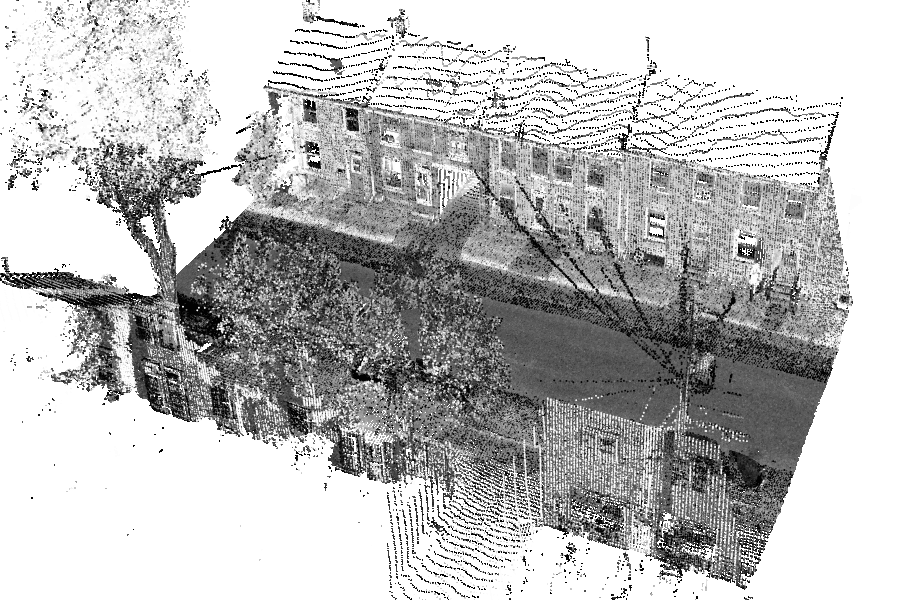}
			\label{fig:donresults_170bagot_orig}
		}
		\subfloat[$\left|\mathbf{\Delta}_\mathbf{\hat{n}}(0.2 \,\textrm{m}, 2 \,\textrm{m})\right|$.]{
			\includegraphics[width=.45\linewidth,keepaspectratio]{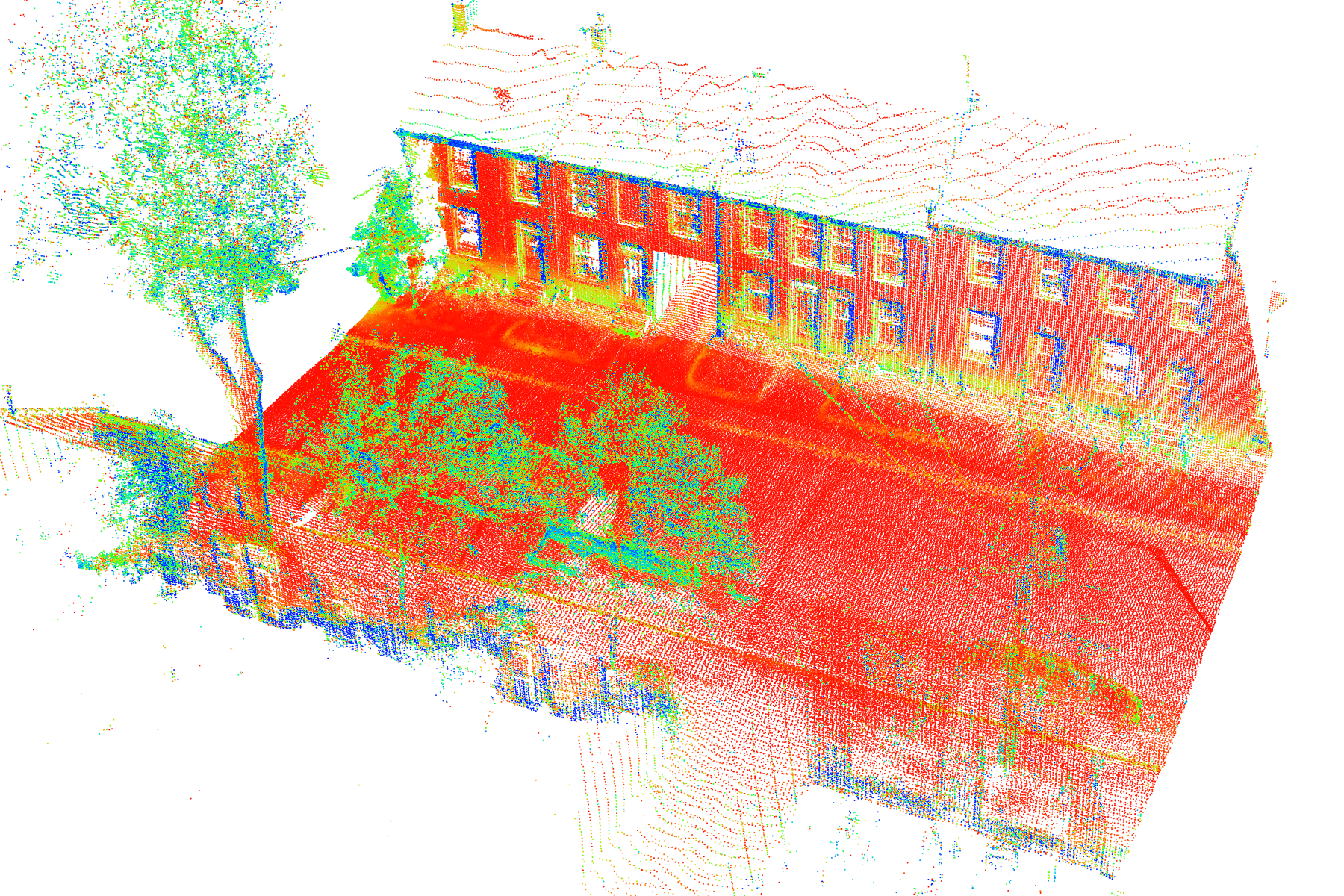}
			\label{fig:donresults_170bagot_0.2}
		}
		\\
		\subfloat[$\left|\mathbf{\Delta}_\mathbf{\hat{n}}(0.8 \,\textrm{m}, 8 \,\textrm{m})\right|$.]{
			\includegraphics[width=.45\linewidth,keepaspectratio]{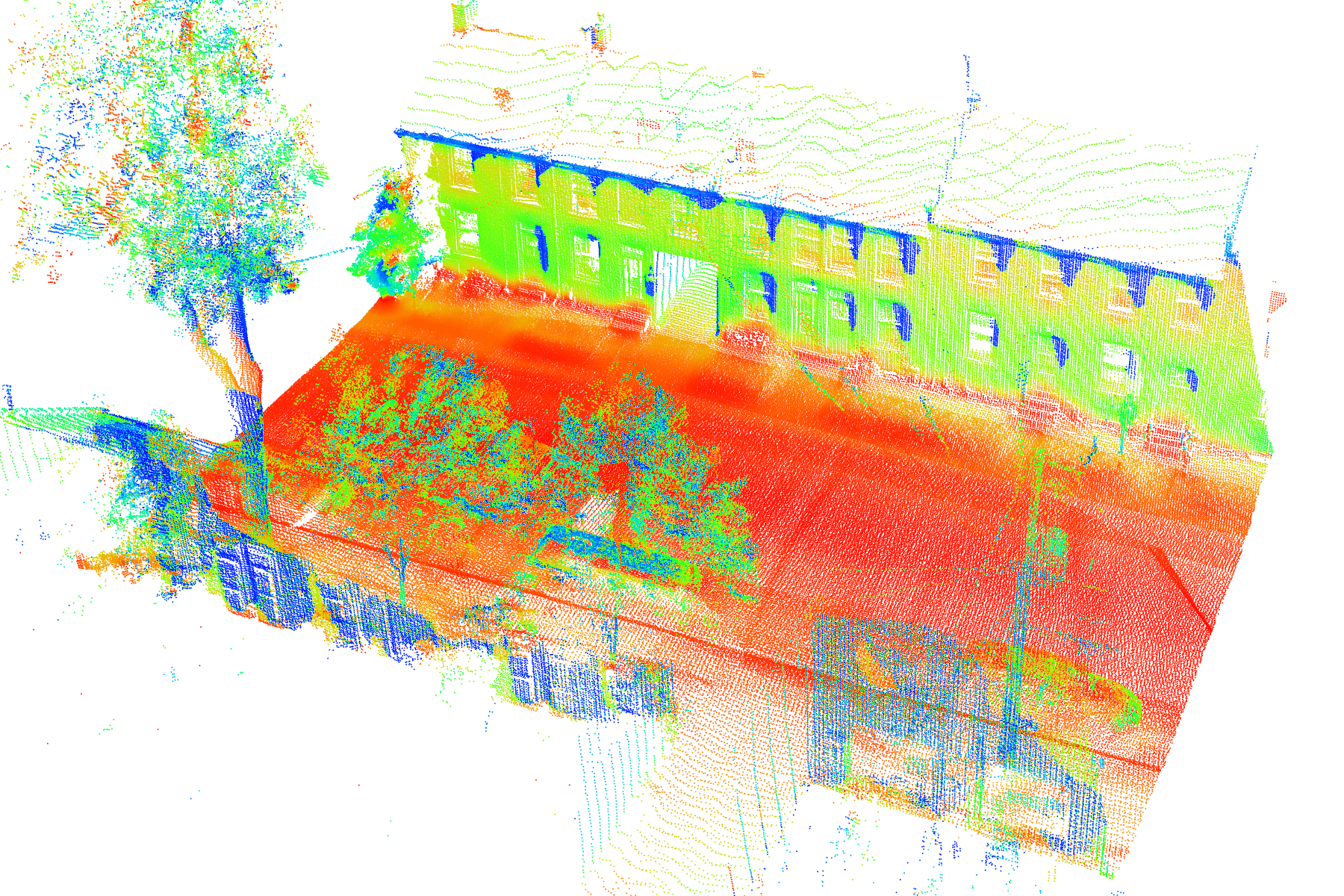}
			\label{fig:donresults_170bagot_0.8}
		}
		\subfloat[$\left|\mathbf{\Delta}_\mathbf{\hat{n}}(2 \,\textrm{m}, 20 \,\textrm{m})\right|$.]{
			\includegraphics[width=.45\linewidth,keepaspectratio]{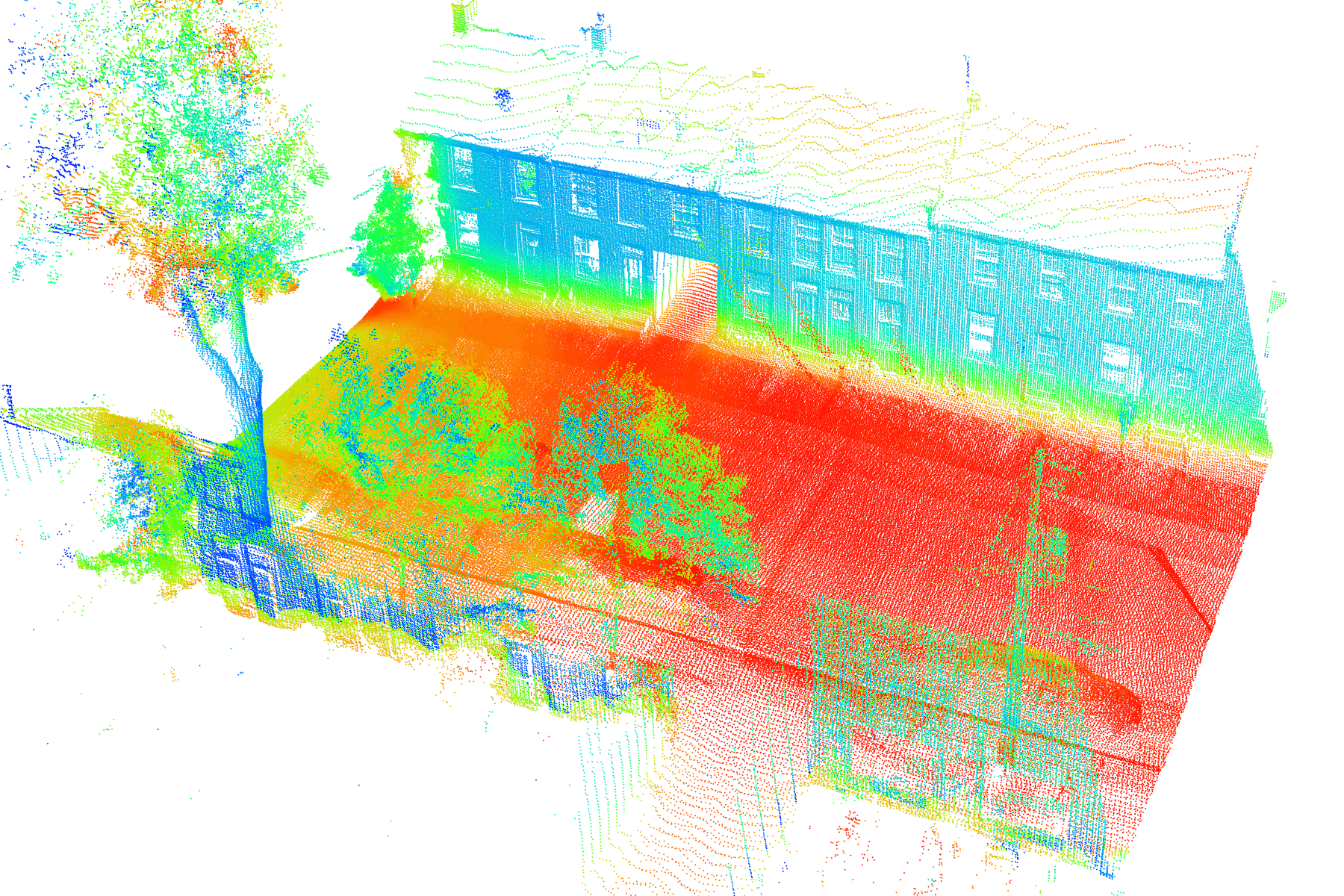}
			\label{fig:donresults_170bagot_2}
		}
	\vspace{-0.5em}
	\subfloat{\includegraphics[width=.65\linewidth,keepaspectratio]{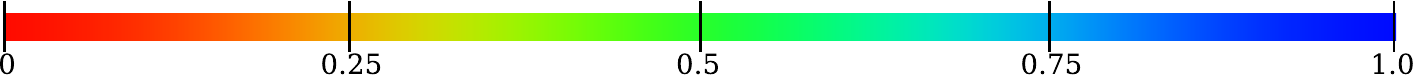}}
	\caption[]{DoN magnitude results on `Bagot St, Kingston, ON, Canada'.}
	\label{fig:donresults_170bagot}
	\end{figure}

	The DoN operator has two parameters, a large radius ($r_2$) and a small radius ($r_1$). While each structure may exhibit a response in a range of scales, it will generally have a natural scale at which this response is maximized. 
	Empirically, it was found that thresholding the magnitude of the $ \mathbf{\Delta}_\mathbf{\hat{n}}(\mathbf{p}) $ vectors obtained with scale ratios ($r_2/r_1$) of $10$ provided good results for filtering out large points belonging to large scale planar surfaces.
\
Fig.~\ref{fig:donresults_170bagot_orig} illustrates a typical urban LIDAR scene, for which the magnitude of the DoN vectors for each scene points (i.e.\ $\|\mathbf{\Delta}_\mathbf{\hat{n}}(\mathbf{p}) \|)$ at three different scales are shown in Figs.~\ref{fig:donresults_170bagot_0.2},~\ref{fig:donresults_170bagot_0.8},~and~\ref{fig:donresults_170bagot_2}. The magnitudes, which are in the $\left[ 0, 1 \right]$ range, are colorized according to the color map shown at the bottom of the image. 

	For DoN parameters corresponding to small scales (e.g.\ within the $0.2-2$~m range), points belonging to lower scale objects have strong responses. For example in Fig.~\ref{fig:donresults_170bagot_0.2}, the finest scale structure exhibits the strongest response. These include road curbs, window ledges, and the details in building facades. 
	For DoN parameters corresponding to larger scales (i.e.\ $2-20$~m), points belonging to larger structures have strong responses. For example in Fig.~\ref{fig:donresults_170bagot_2} the building points have a large response, yet very large scale structures (i.e.\ the road surface) still exhibits a small response. 

	\subsubsection{DoN Scale-Based Filtering}
	\label{subsec:filtering}
		\begin{figure}[htbp]
		\centering
		\subfloat[Original: 614403 points.]{
			\includegraphics[width=.45\linewidth,keepaspectratio]{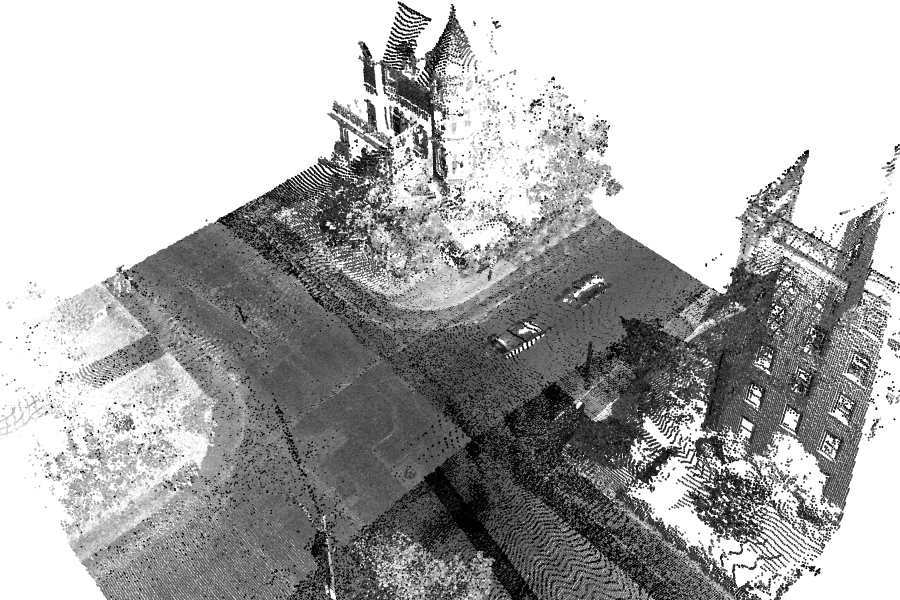}
			\label{fig:donsegresults_clergyandjohnson_orig}
		}
		\subfloat[$\left|\mathbf{\Delta}_\mathbf{\hat{n}}(0.1 \,\textrm{m}, 1 \,\textrm{m})\right|\geq 0.25$: 135518 points.]{
			\includegraphics[width=.45\linewidth,keepaspectratio]{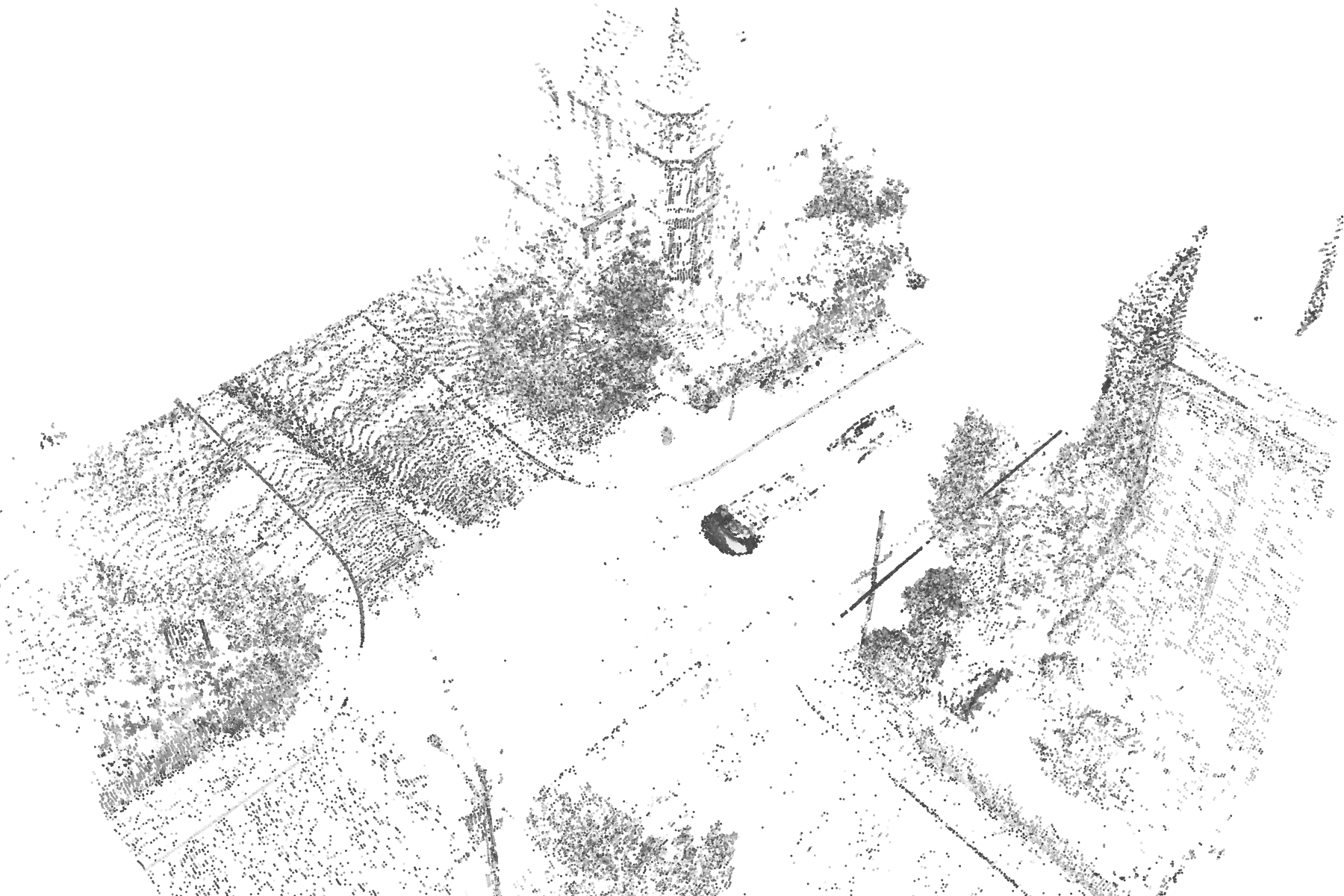}
			\label{fig:donsegresults_clergyandjohnson_0.1}
		}
		\\
		\subfloat[$\left|\mathbf{\Delta}_\mathbf{\hat{n}}(0.2 \,\textrm{m}, 2 \,\textrm{m})\right|\geq 0.25$: 132708 points.]{
			\includegraphics[width=.45\linewidth,keepaspectratio]{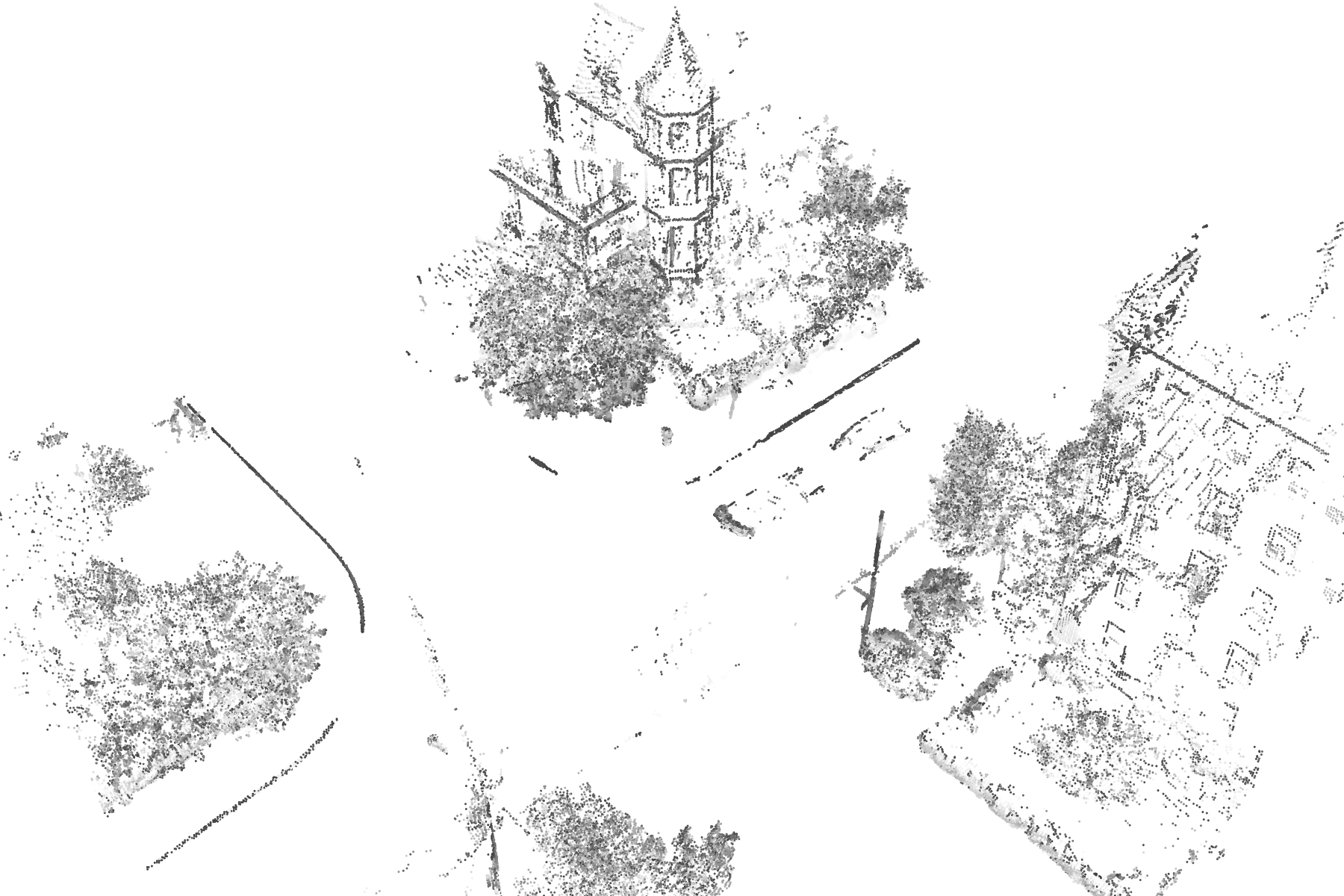}
			\label{fig:donsegresults_clergyandjohnson_0.2}
		}
		\subfloat[$\left|\mathbf{\Delta}_\mathbf{\hat{n}}(0.8 \,\textrm{m}, 8 \,\textrm{m})\right|\geq 0.25$: 139367 points.]{
			\includegraphics[width=.45\linewidth,keepaspectratio]{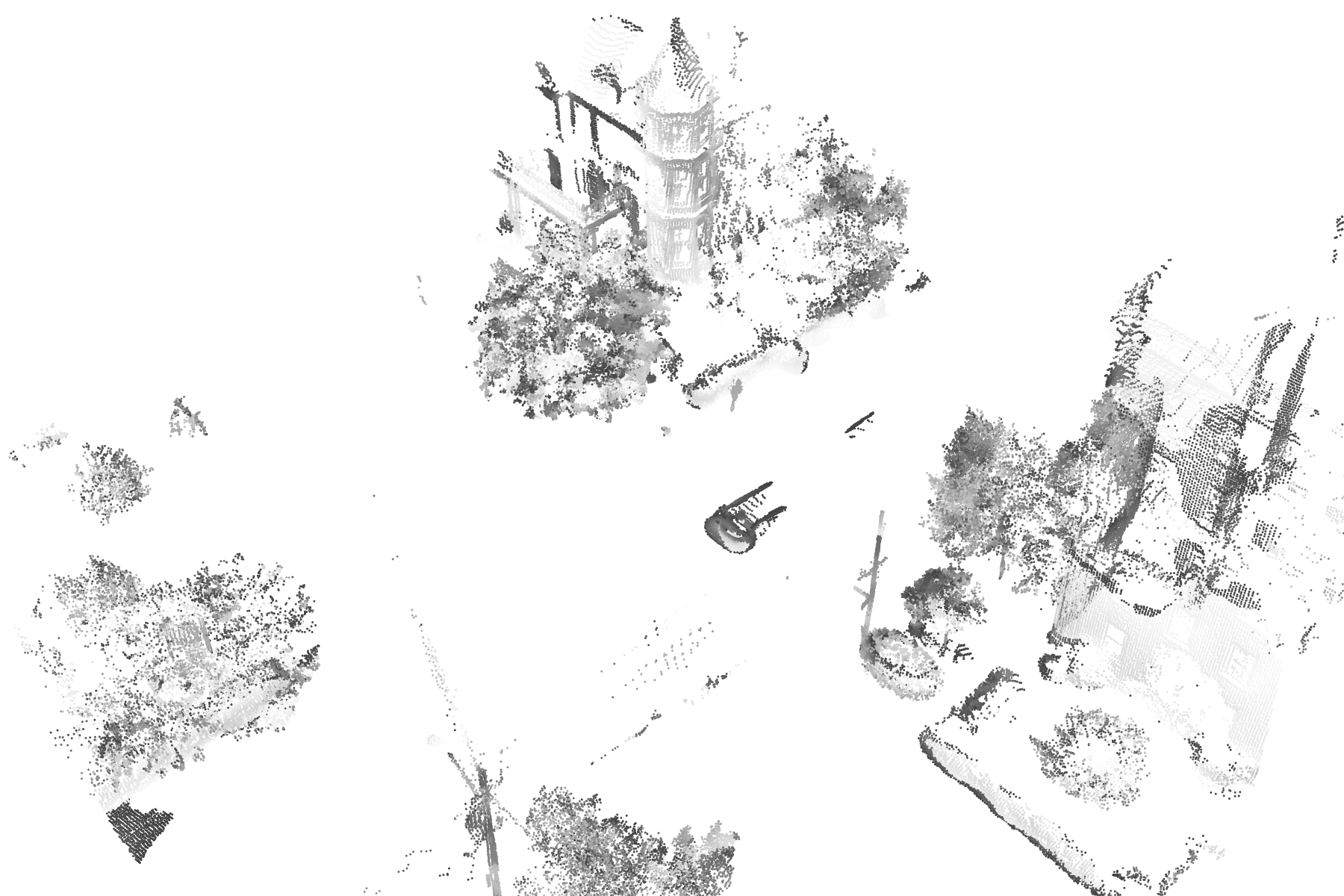}
			\label{fig:donsegresults_clergyandjohnson_0.8}
		}
	\caption[]{DoN filtering results on `Intersection of Clergy and Johnson St, Kingston, ON, Canada'.}
	\label{fig:donsegresults_clergyandjohnson}
	\end{figure}

	An important application of the DoN operator, as motivated by the results shown in Fig.~\ref{fig:donresults_170bagot}, is to use it as a salience operator to pre-filter point clouds. Fig.~\ref{fig:donsegresults_clergyandjohnson} shows the results of such a filtering of a point cloud, discarding all the points for which $\left\|\mathbf{\Delta}_\mathbf{\hat{n}}(\mathbf{p})\right\| < 0.25 $, on a typical urban scene, with various DoN parameters corresponding to a range of scales. 

	At the lowest scale ($0.1-1.0$~m), shown in Fig.~\ref{fig:donsegresults_clergyandjohnson_0.1}, sharp edges are clearly preserved, including building edges (window outlines and pipes) and ground edges (street curbs). Also preserved, however, is artificial structure derived from noise - to be expected at approximately the resolution of the data. By the next, incrementally larger, scale ($0.2-2.0$~m), shown in Fig.~\ref{fig:donsegresults_clergyandjohnson_0.2}, the noise has been filtered out. As the scale is increased, larger and larger objects are preserved, while smaller objects are increasingly discarded. At the largest scale ($0.8-8.0$~m), shown  in Fig.~\ref{fig:donsegresults_clergyandjohnson_0.8}, larger building fronts and walls are segmented from the rest of the scene.

	\subsubsection{Segmentation}
	\label{subsec:segmentation}
		\begin{figure*}[htbp]
		\centering
			\subfloat[Original point cloud (620,820 points).]{
				\includegraphics[width=.45\linewidth,keepaspectratio]{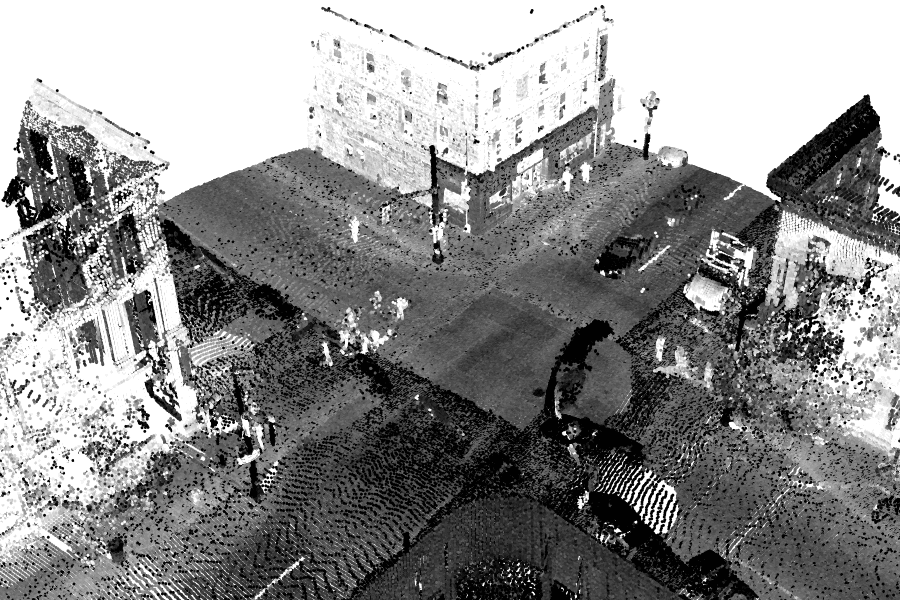}
				\label{fig:donclusterresults_princessandbagot_orig}
			}
			\subfloat[Clusters found in $\left|\mathbf{\Delta}_\mathbf{\hat{n}}(0.2 \,\textrm{m}, 2 \,\textrm{m})\right|\geq 0.25$.]{
				\includegraphics[width=.45\linewidth,keepaspectratio]{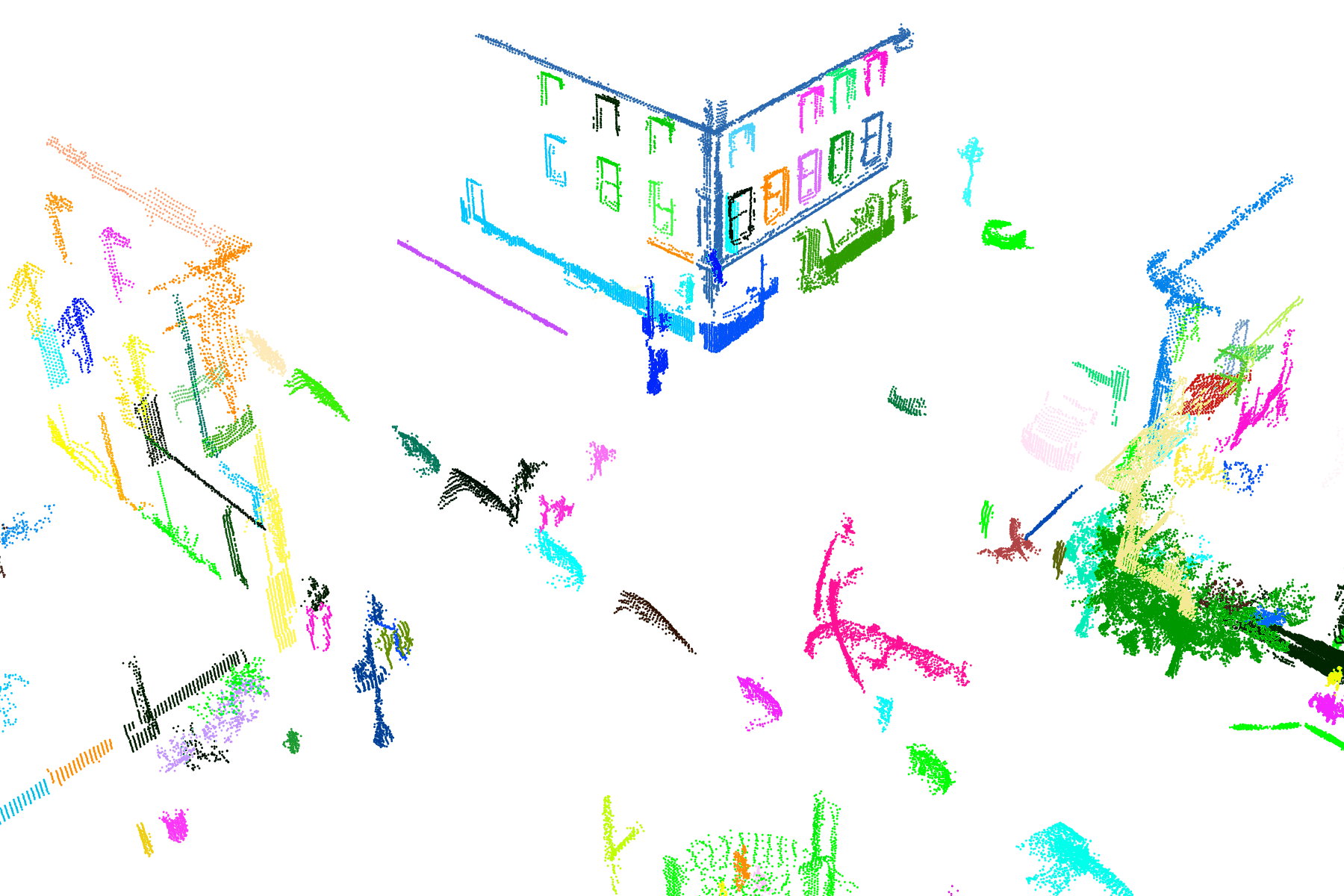}
				\label{fig:donclusterresults_princessandbagot_0.2}
			}
\\
			\subfloat[Person cluster.]{
				\includegraphics[width=.15\linewidth,keepaspectratio]{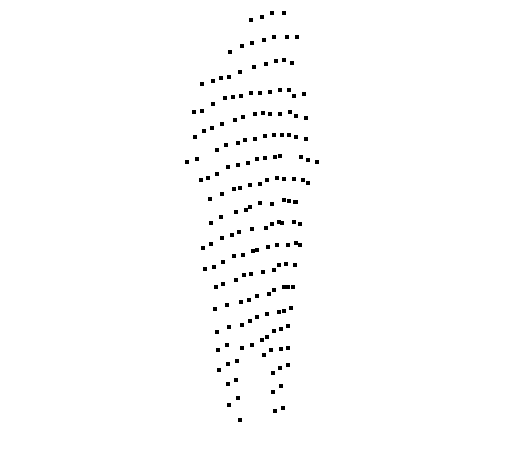}
				\label{fig:donclusterresults_princessandbagot_0.2_person}
			}
			\subfloat[Traffic light cluster.]{
				\includegraphics[width=.15\linewidth,keepaspectratio]{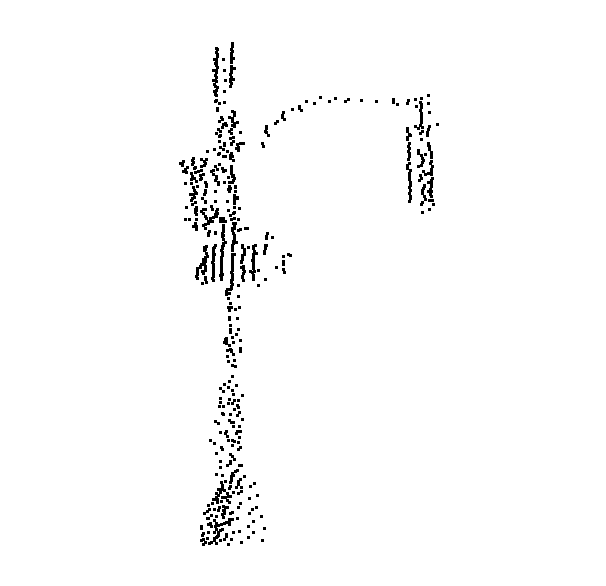}
				\label{fig:donclusterresults_princessandbagot_0.2_trafficlight}
			}
			\subfloat[Window cluster.]{
				\includegraphics[width=.15\linewidth,keepaspectratio]{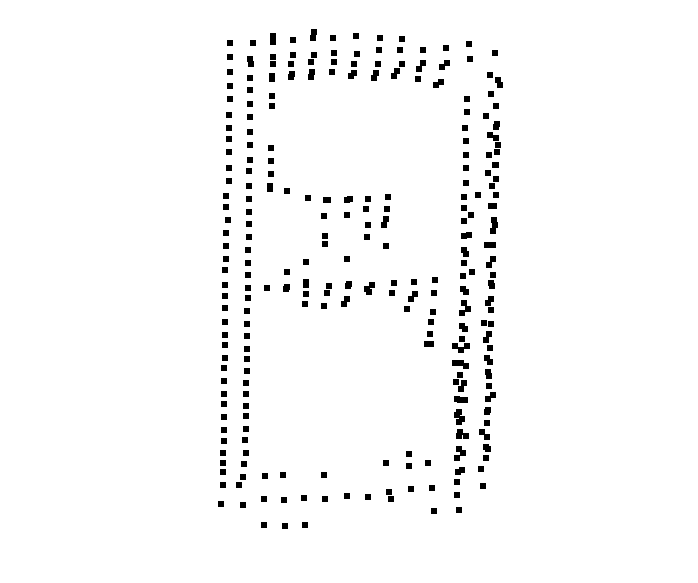}
				\label{fig:donclusterresults_princessandbagot_0.2_window}
			}
			\subfloat[Car cluster.]{
				\includegraphics[width=.15\linewidth,keepaspectratio]{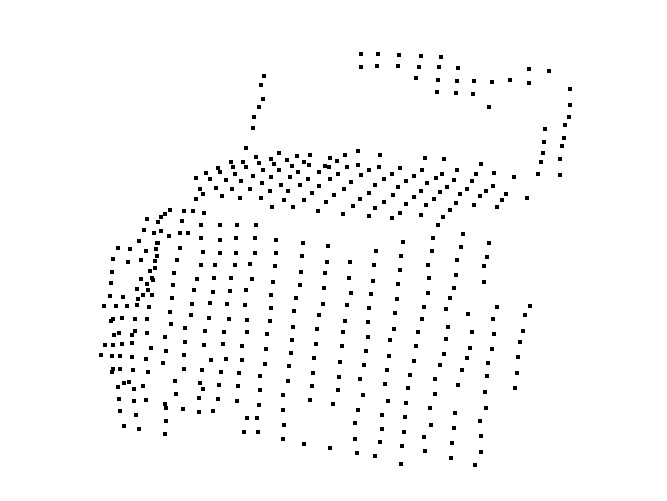}
				\label{fig:donclusterresults_princessandbagot_0.2_car}
			}
			\subfloat[Tree cluster.]{
				\includegraphics[width=.15\linewidth,keepaspectratio]{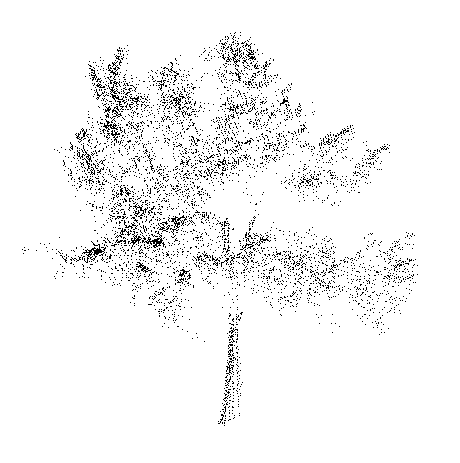}
				\label{fig:donclusterresults_princessandbagot_0.2_tree}
			}
	\caption[]{DoN clustering results for `Intersection of Princess and Bagot St, Kingston, ON, Canada' and sample clusters from.}
	\label{fig:donclusterresults_princessandbagot}
	\end{figure*}

	DoN filtering of a point cloud, such as that described in \S\ref{subsec:filtering}, was found to result in good isolation of points in urban LIDAR scenes. Applying a simple clustering method to the resulting point cloud, results in the clear clustering of many objects of interest in a scene. A simple Euclidean distance threshold based clustering algorithm, (Euclidean Cluster Extraction~\cite{RusuDoctoralDissertation}), was applied with a distance tolerance of $r_1$, a minimum of $100$ cluster points, and a maximum of $100,000$ cluster points.	
	Fig.~\ref{fig:donclusterresults_princessandbagot} shows the results of such clustering to DoN filtered scenes of various DoN parameters. Each cluster in the scene is assigned a random (non-unique) color. 
	Figs~\ref{fig:donclusterresults_princessandbagot_0.2_person}-\ref{fig:donclusterresults_princessandbagot_0.2_tree} illustrate various clusters in the scene, corresponding to various objects including a person, a traffic light fixture, a window, a car, and a tree.
	
	Such segmentation might form a fundamental pre-processing step in an object recognition pipeline for finding objects in an urban LIDAR scene~\cite{taati}. The pipeline would include the described clustering method, followed by feeding individual clusters into an object recognition algorithm. Since the the clustering algorithm isolates individual objects (such as cars, people, fire hydrants, etc), DoN clustering enables the use of global object recognition methods that require pre-segmentation~\cite{liminicv}.

\subsection{KITTI Vision Benchmark Suite}
\label{sec:kittiresults}
\subsubsection{KITTI Dataset}
\begin{figure}[htb]
	\centering
        \includegraphics[width=0.95\linewidth,keepaspectratio]{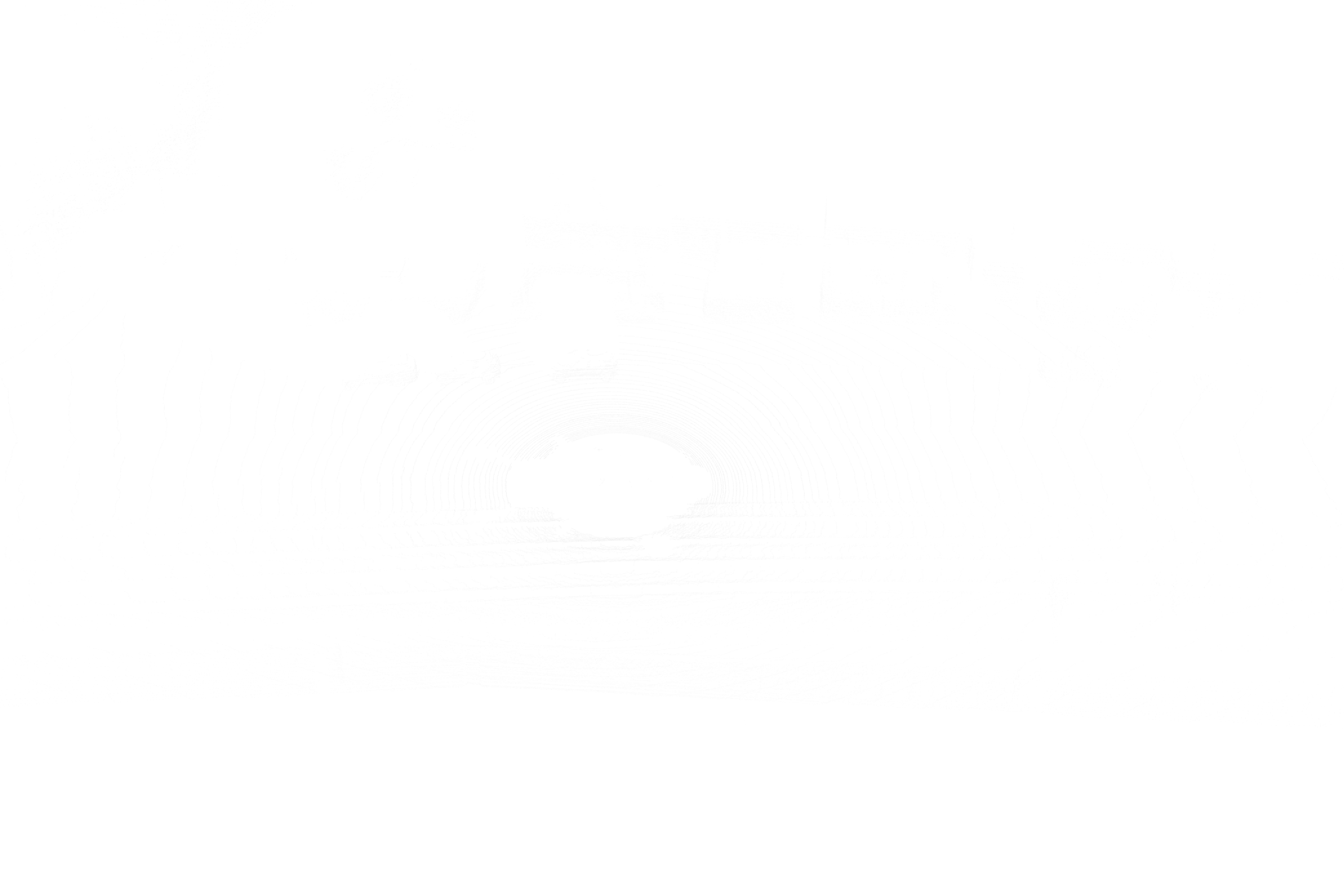}
	\caption[]{A single frame KITTI Velodyne point cloud.}
        \label{fig:kittidata}
	\vspace{-0.5em}
\end{figure}

	Although the TITAN Urban Mobile LIDAR Data application was motivational in the initial development of the proposed method, it is not a public dataset and does not have readily available ground truth. Thus for a repeatable quantitative evaluation, we used the KITTI Vision Benchmark Suite\cite{KITTI}. The KITTI dataset includes a large number of point clouds along with annotated ground truth bounding boxes for objects of interest to driving and navigation. Each sequence consists of a number of frames, where each frame has an inertially corrected 3D Velodyne point cloud ($\sim$100k points per frame), and manually annotated 3D object bounding boxes for cars, trucks, trams, pedestrians, and cyclists.

	Although the KITTI data also consists of unorganized point clouds, it is far sparser than the TITAN point clouds, and was captured with a single $360\,^{\circ}$ sensor rather than an array of line scanners. Fig.~\ref{fig:kittidata} illustrates a sample Velodyne point cloud from a single frame. 
	\subsubsection{Method}
	In order to evaluate DoN based segmentation, as illustrated in \S\ref{subsec:segmentation}, a set of DoN parameters ($r_1, r_2$) and DoN magnitude thresholds $t$ were chosen based on the parameter selection algorithm outlined in \S\ref{sec:parameterselection}. As in the method outlined through sections \S\ref{subsec:donoutput} through \S\ref{subsec:segmentation}, the DoN was calculated for a sequence of frames (point clouds) after which the DoN magnitude was thresholded by a fixed value ($t = 0.25$) and Euclidean Cluster Extraction~\cite{PCL} was performed with a distance threshold equivalent to the smallest DoN radius $r_1$ and a set of clusters extracted. 

	For each frame, the set of clusters was compared with the each of the ground truth bounding box labels to identify the cluster with highest intersection. This candidate cluster was then compared with the ground truth point cloud by collecting various statistics. 

	The main measures used to evaluate quality of the segmentation were precision (ratio of correctly predicted object points to the total number of predicted object points), and recall (ratio of correctly predicted object points to the number of ground truth object points). 

	Due to the nature of the Velodyne data, in many frames the point clouds within ground truth bounding boxes may consist of very few points ($<100$). It was judged that such extremely sparse ground truth objects were unsuitable for evaluating the segmentation of smaller scale objects, and since our clustering algorithm's minimum threshold was 100 points, for all of the object classes a minimum of 100 points was required for a ground truth point cloud to be used in evaluation.	
	\subsubsection{KITTI Results}	
\begin{figure}[htbp]
	\centering
	\subfloat{
	\centering
\hspace{3em}
$\vcenter{\hbox{\includesvg{Pedestriankey}\hspace{0.1em}}}$
$\vcenter{\hbox{\includesvg{Cyclistkey}\hspace{0.3em}}}$
$\vcenter{\hbox{\includesvg{Carkey}\hspace{0.1em}}}$
$\vcenter{\hbox{\includesvg{Vankey}\hspace{0.1em}}}$
$\vcenter{\hbox{\includesvg{Truckkey}}}$}\\
   	\vspace{-0.5em}
    \subfloat[Clusters from $\Delta_{\hat{n}}(0.1, 1.4) \ge 0.25$ with threshold $0.2$]{
        	\includegraphics[width=1\linewidth,keepaspectratio]{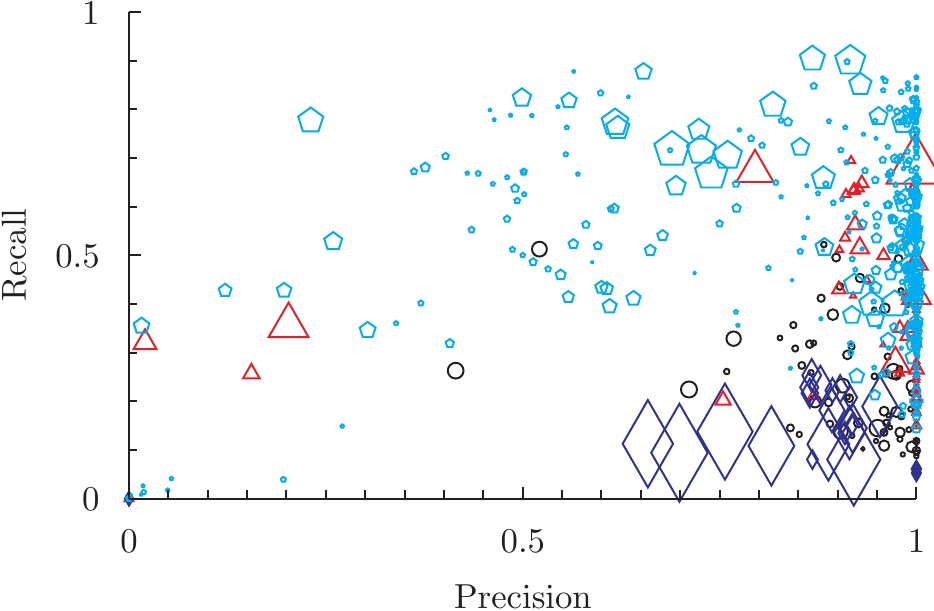}
                \label{fig:kitti_0.1_1.4_results}
        }\\
        \vspace{-0.5em}
	\subfloat[Clusters from $\Delta_{\hat{n}}(0.2, 2) \ge 0.25$ with threshold $0.2$]{
        	\includegraphics[width=1\linewidth,keepaspectratio]{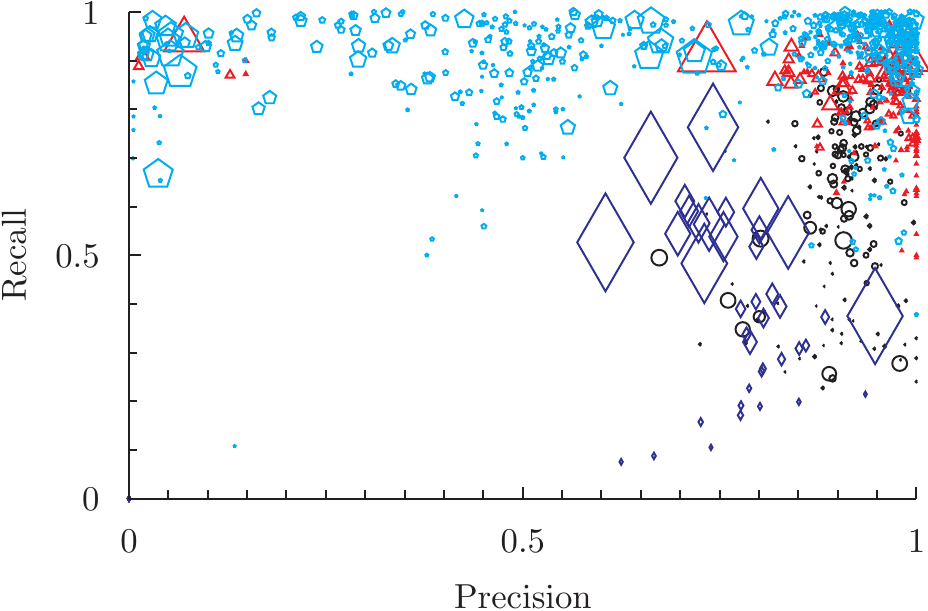}
                \label{fig:kitti_0.2_2_results}
        }\\
        \vspace{-0.5em}
	\subfloat[Clusters from $\Delta_{\hat{n}}(1, 3) \ge 0.25$ with threshold $1.0$]{
        	\includegraphics[width=1\linewidth,keepaspectratio]{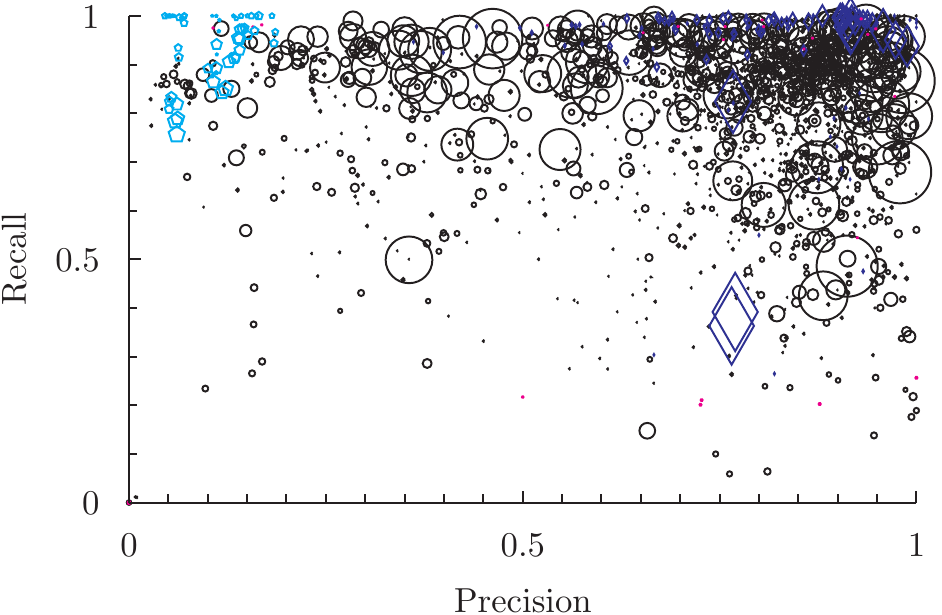}
                \label{fig:kitti_1_3_results}
        }
        \vspace{-0.5em}
\caption[]{Results of DoN clustering v.s. ground truth.}
        \label{fig:kittiresults}
\end{figure}
	Fig.~\ref{fig:kittiresults} illustrates the results of our evaluation in the form of a precision/recall graph over thousands of ground truth objects on two different sequences in the KITTI dataset, \verb|2011_09_26_drive_0001| (Fig.~\ref{fig:kitti_0.1_1.4_results},\ \ref{fig:kitti_0.2_2_results}) and \verb|2011_09_26_drive_0009| (Fig.~\ref{fig:kitti_1_3_results}). Each data point is of a size proportional to it's ground truth point cloud's size (note: scale is not preserved inter-class). 

	The majority of the results have a $\mathrm{precision}>0.9$. However the recall values depend more on the class of object and DoN parameters. Smaller scale objects, such as pedestrians and cyclists have higher recall/precision for the smaller parameters of $r_1, r_2$ as can be seen in Fig.~\ref{fig:kitti_0.1_1.4_results},\ \ref{fig:kitti_0.2_2_results}. While larger scale objects such as cars and vans have higher recall/precision for larger radii, as can be seen in Fig.~\ref{fig:kitti_1_3_results}. 

	While is is difficult to compare the performance of algorithms evaluated on different tests sets (we advocate the usage of the public KITTI dataset), the results appear favourable in comparison with the recall/precision of more computationally intensive mesh and graph-based segmentation methods evaluated on less challenging datasets~\cite{Golovinskiy_Funkhouser_2009}.
\subsection{Computational Efficiency}
	\label{subsec:computation}
	The computation of DoN on a scene requires the calculation of the normal maps and is bound by the nearest neighbor radius search for the largest radius parameter $r_2$. In practice, for large radii, this can involve calculating the normal to a point using hundreds of thousands of points. Instead an approximation can be calculated by uniformly sub-sampling the point cloud used for the nearest neighbor search. The current implementation of DoN can do this given a decimation parameter $d$, where the search point cloud is sub-sampled using a uniform re-sampling algorithm, with the point cloud coarsely voxelized into voxels of length $r_1/d$ for the small radius normals and $r_2/d$ for the large radius normals. It was found that an approximation with $d=10$ results in negligible error, while halving the time for calculating DoN with $r_1=0.1, r_2=1.0$ on the point cloud in Fig.~\ref{fig:donresults_170bagot} with 478,348 points to 3454.33~ms compared with 7812.5~ms for the full calculation on a 3.2Ghz i7. A preliminary GPU implementation of DoN was found to be an order of magnitude faster, taking only 565.46 ms on an NVIDIA GTX 480 for the full computation.

\section{Conclusion and Future Work}
\label{sec:conclusion}
The Difference of Normals as a multi-scale operator was introduced for unorganized 3D point clouds. Illustrated results on dense urban LIDAR data qualitatively showcased the effectiveness of DoN filtering in keeping points belonging to objects at a given scale, while discarding those belonging to structures of other scales. The application of DoN as a scale-base filtering and segmentation tool was highlighted in urban LIDAR scenes. Results on a typical urban street intersection with clustering showed a clear segmentation of points belonging to various objects of interest at different scales, such as cars, road curbs, trees, and buildings - some having as few as 100 points. With urban LIDAR scenes typically containing millions of points, DoN filtering provides a substantial reduction in points for performing any further processing of the scene.

The quality of DoN-based segmentation was quantitatively evaluated on a large, publicly available dataset of sparse, unorganized urban LIDAR data. Objects such as cars and pedestrians were automatically segmented from the scene and compared with ground truth annotations.

Future work includes the development of a DoN based surface descriptor to exploit the defined scale operator over several radii, and integration with object recognition methods. The development of an interactive semi-automated tool for annotating large scale 3D point clouds in particular would go a long way towards simplifying the generation of GIS models from urban LIDAR point clouds. 

\section*{Acknowledgments}
The authors thank Geodigital Inc.\ for their support and providing us access to the TITAN data. We also acknowledge the financial support of the Geoide Network Center of Excellence of Natural Sciences and Engineering Research Council of Canada, and the Ontario Centres of Excellence.

{\small
\bibliographystyle{ieee}
\bibliography{thesis}

\begin{thebibliography}{10}\itemsep=-1pt

\bibitem{alexanormals}
M.~Alexa, J.~Behr, D.~Cohen-Or, S.~Fleishman, D.~Levin, and C.~T.~Silva.
\newblock Computing and rendering point set surfaces.
\newblock {\em IEEE Transactions on Visualization and Computer Graphics},
  9(1):3--15, 2003.

\bibitem{KITTI}
A.~Geiger, P.~Lenz, and R.~Urtasun.
\newblock Are we ready for autonomous driving? the kitti vision benchmark
  suite.
\newblock In {\em Computer Vision and Pattern Recognition (CVPR)}, Providence,
  USA, June 2012.

\bibitem{titan}
C.~Glennie.
\newblock Reign of point clouds: A kinematic terrestrial lidar scanning system.
\newblock {\em Inside GNSS}, (Fall), 2007.

\bibitem{Golovinskiy_Funkhouser_2009}
A.~Golovinskiy and T.~Funkhouser.
\newblock Min-cut based segmentation of point clouds.
\newblock {\em 2009 IEEE 12th International Conference on Computer Vision
  Workshops ICCV Workshops}, 150:39--46, 2009.

\bibitem{hoppenormals}
H.~Hoppe, T.~DeRose, T.~Duchamp, J.~McDonald, and W.~Stuetzle.
\newblock Surface reconstruction from unorganized points.
\newblock In {\em SIGGRAPH {'}92: Proceedings of the 19th annual conference on
  Computer graphics and interactive techniques}, pages 71--78, New York, NY,
  USA, 1992. ACM.

\bibitem{huinormals}
H.~Huang, D.~Li, H.~Zhang, U.~Ascher, and D.~Cohen-Or.
\newblock Consolidation of unorganized point clouds for surface reconstruction.
\newblock {\em ACM Trans. Graph.}, 28(5):1--7, 2009.

\bibitem{Huang}
J.~Huang and C.-H. Menq.
\newblock Automatic data segmentation for geometric feature extraction from
  unorganized 3-d coordinate points.
\newblock {\em Robotics and Automation, IEEE Transactions on}, 17(3):268--279,
  Jun 2001.

\bibitem{koenderink}
J.~J. Koenderink.
\newblock The structure of images.
\newblock {\em Biological Cybernetics}, 50:363--370, 1984.

\bibitem{robustnormals}
B.~Li, R.~Schnabel, R.~Klein, Z.~Cheng, G.~Dang, and J.~Shiyao.
\newblock Robust normal estimation for point clouds with sharp features.
\newblock {\em Computers {\&} Graphics}, 34(2):94--106, Apr. 2010.

\bibitem{scalespacebook}
T.~Lindeberg.
\newblock Scale-space theory: A basic tool for analysing structures at
  different scales.
\newblock {\em Journal of Applied Statistics}, 21:224--270, 1994.

\bibitem{cms}
Y.~Liu and Y.~Xiong.
\newblock Automatic segmentation of unorganized noisy point clouds based on the
  gaussian map.
\newblock {\em Computer-Aided Design}, 40(5):576--594, 2008.

\bibitem{SIFT}
D.~G. Lowe.
\newblock Distinctive image features from scale-invariant keypoints.
\newblock {\em International Journal of Computer Vision}, 60(2):91--110, 2004.

\bibitem{maar_hildreth}
Maar and Hildreth.
\newblock Theory of edge detection.
\newblock In {\em Proceedings of Royal Society of London}, pages 187--217,
  1980.

\bibitem{RusuDoctoralDissertation}
R.~B. Rusu.
\newblock {\em Semantic 3D Object Maps for Everyday Manipulation in Human
  Living Environments}.
\newblock PhD thesis, Computer Science department, Technische Universitaet
  Muenchen, Germany, October 2009.

\bibitem{PCL}
R.~B. Rusu and S.~Cousins.
\newblock {3D is here: Point Cloud Library (PCL)}.
\newblock In {\em {IEEE International Conference on Robotics and Automation
  (ICRA)}}, Shanghai, China, May 9-13 2011.

\bibitem{Rusu_Marton_Blodow_Beetz_2008}
R.~B. Rusu, Z.~C. Marton, N.~Blodow, and M.~Beetz.
\newblock Persistent point feature histograms for 3d point clouds.
\newblock {\em autonomous systems 10}, page 119, 2008.

\bibitem{liminicv}
L.~Shang and M.~Greenspan.
\newblock Real-time object recognition in sparse range images using error
  surface embedding.
\newblock {\em International Journal of Computer Vision.}, 89(2-3), 2010.

\bibitem{taati}
B.~Taati and M.~Greenspan.
\newblock Local shape descriptor selection for object recognition in range
  data.
\newblock {\em Computer Vision and Image Understanding}, 115(5):681--694, 2011.

\bibitem{Unnikrishnan_2008}
R.~Unnikrishnan and M.~Hebert.
\newblock Multi-scale interest regions from unorganized point clouds.
\newblock {\em Computer Vision and Pattern Recognition Workshop}, 0:1--8, 2008.

\bibitem{witkin}
A.~Witkin.
\newblock Scale-space filtering: A new approach to multi-scale description.
\newblock In {\em Acoustics, Speech, and Signal Processing, IEEE International
  Conference on ICASSP {'}84.}, volume~9, pages 150--153, Mar 1984.

\bibitem{woo}
H.~Woo, E.~Kang, S.~Wang, and K.~H. Lee.
\newblock A new segmentation method for point cloud data.
\newblock {\em International Journal of Machine Tools and Manufacture},
  42(2):167--178, 2002.

\bibitem{volumetricscale}
C.~Wyatt, E.~Bayram, and Y.~Ge.
\newblock Minimum reliable scale selection in 3d.
\newblock {\em Pattern Analysis and Machine Intelligence, IEEE Transactions
  on}, 28(3):481--487, March 2006.

\end{thebibliography}
}
\end{document}